\documentclass[11pt,letterpaper]{article}

\usepackage{acl}

\usepackage{times}
\usepackage{latexsym}
\usepackage[T1]{fontenc}
\usepackage[utf8]{inputenc}
\usepackage{microtype}
\usepackage{inconsolata}
\usepackage{graphicx}

\usepackage{booktabs}
\usepackage{amssymb}
\usepackage{enumitem}
\usepackage{array}

\title{Memory Architectures for Multi-Turn Text-to-SQL:\\A Benchmark and Empirical Study}

\author{Ravi Kumar Tummalapenta \quad Suman Addanki \\
  LLM Suite Engineering Team, JP Morgan Chase \& Co. \\
  Email: \texttt{\{ravi.k.tummalapenta, suman.k.addanki\}@jpmchase.com}}

\begin{document}
\maketitle
\sloppy

\begin{abstract}
Multi-turn Text-to-SQL---where analysts refine structured database queries across conversational turns---is central to enterprise AI deployments, yet Text-to-SQL systems remain predominantly evaluated in single-turn settings. We ask three questions about this gap. First, do frontier large language models exhibit measurable degradation in multi-turn Text-to-SQL when operating without explicit memory support? Second, among common memory architectures---working, episodic, and semantic---which components contribute meaningfully to multi-turn performance and at what marginal cost? Third, do model families differ in how they utilize memory, and can such differences be attributed to capability gaps or to evaluation asymmetries?

To answer these questions we introduce \textsc{EnterpriseMem-Bench}, a multi-turn Text-to-SQL benchmark comprising 300 conversational sessions and 1,400 turns constructed programmatically from three enterprise domains: the BIRD financial database, SEC EDGAR quarterly filings for twenty Fortune 500 companies, and the Northwind enterprise sales schema. Each session is generated deterministically from live database profiles with ground-truth SQL authored as parameterized templates verified against source databases; each turn is annotated with a deterministic memory-critical flag indicating whether correct SQL generation requires information introduced in a prior turn. We evaluate five frontier models---GPT-5 mini, GPT-5.2, Claude Sonnet 4.5, Claude Sonnet 4.6, and Claude Opus 4.6---across five memory conditions whose design enables a three-way ablation isolating working-memory window size, episodic retrieval, and semantic augmentation as independent effects. All Claude models are evaluated with extended thinking enabled to maintain methodological parity with GPT reasoning models, which reason by default. We introduce the Memory Benefit Score (MBS) as a per-turn diagnostic metric normalized against the stateless baseline.

Our experiments yield four principal findings. First, stateless multi-turn Text-to-SQL collapses universally: all five models drop to zero execution accuracy by the third turn on memory-critical questions, even under extended reasoning. Second, the largest single architectural gain comes from a two-turn working-memory window; episodic and semantic layers provide model- and dataset-dependent marginal effects that are frequently negative, indicating that memory-architecture complexity does not monotonically improve accuracy. Third, a reproducible, model-specific regression is observed on SEC EDGAR financial queries, where Claude Sonnet 4.6 underperforms both its predecessor (Sonnet 4.5) and its sibling (Opus 4.6) by $17$--$33$ percentage points across memory conditions---a finding that survives extended reasoning. Fourth, under reasoning, Claude error distributions become mono-modal: preamble, syntax, and runtime errors are eliminated, and every remaining failure is a wrong-result error. We release \textsc{EnterpriseMem-Bench} together with the session generation pipeline, agent implementation, and evaluation harness to support reproducibility and downstream work.
\end{abstract}

\renewcommand{\thefootnote}{\fnsymbol{footnote}}
\footnotetext[1]{\scriptsize This paper was prepared for informational purposes by the LLM Suite group of JP Morgan Chase \& Co. and its affiliates (``JPMC'') and is not a product of the Research Department of JP Morgan. JP Morgan makes no representation, warranty or undertaking whatsoever and disclaims all liability for the completeness, accuracy or reliability of the information contained herein. This document is not intended as investment research or investment advice, or a recommendation, offer or solicitation for the purchase or sale of any security, financial instrument, financial product or service, or to be used in any way for evaluating the merits of participating in any transaction, and shall not constitute a solicitation under any jurisdiction or to any person, if such solicitation under such jurisdiction or to such person would be unlawful.}
\renewcommand{\thefootnote}{\arabic{footnote}}

\section{Introduction}

An equity research analyst opens an internal SQL assistant and asks: ``Show me Apple's revenue for 2022 and 2023.'' The system generates valid SQL, executes it, and returns two rows. She reads the result and asks a follow-up: ``Which year was higher?'' The second question contains no entity, no metric, no filter of its own---it refers entirely to information from the prior turn. A system that treats each question independently cannot answer it. A system that retains the prior turn answers trivially.

This scenario is representative rather than exceptional. Enterprise data-analytics conversations are overwhelmingly multi-turn: follow-up questions form the core of analyst workflows, and each follow-up typically carries anaphoric references (``that company''), ellipsis (``and 2024?''), or implicit continuations of prior filters. Yet the overwhelming majority of Text-to-SQL research and benchmarking evaluates models in single-turn settings, where each question is independently complete and memory plays no role. When multi-turn performance is reported, it is typically as an aggregate accuracy number across conversational data; the specific architectural question of what kind of memory a Text-to-SQL system needs, and at what cost, has been largely unexamined.

This paper investigates that gap through a controlled empirical study. We construct a benchmark designed specifically for memory evaluation---not merely a multi-turn benchmark, but one whose sessions are instrumented such that each turn carries a deterministic flag indicating whether answering it requires information introduced in a prior turn. We define five memory conditions of increasing architectural complexity, designed so that three of the conditions form a clean independent-effects ablation: we can isolate the contribution of working-memory window size from the contribution of episodic retrieval, and isolate both from the contribution of semantic-schema augmentation. We evaluate five frontier large language models spanning two providers and multiple generations. We introduce a per-turn diagnostic metric, the Memory Benefit Score, that normalizes each condition against its stateless baseline.

The study is organized around three research questions:

\paragraph{RQ1.} When operating without explicit memory support, do frontier large language models exhibit measurable degradation in multi-turn Text-to-SQL? If so, how severe is the degradation, and does it depend on model family, reasoning configuration, or dataset?

\paragraph{RQ2.} Among standard memory architectures---working memory, episodic retrieval, and semantic augmentation---which components contribute meaningfully to multi-turn performance? Does memory-architecture complexity monotonically improve accuracy, or are there diminishing or negative returns?

\paragraph{RQ3.} Do model families differ systematically in how they utilize memory? Can observed differences be attributed to capability gaps between models, or to evaluation asymmetries introduced by differences in how reasoning is handled across providers?

These questions dictate the study's design. RQ1 requires a benchmark in which the stateless condition is well-defined and memory-critical turns are identifiable, so that stateless-baseline collapse can be measured rather than assumed. RQ2 requires a memory-architecture space that permits clean ablation of individual components, which rules out simply comparing ``full memory'' against ``no memory'' and demands a design where window size, retrieval, and augmentation can be isolated from one another. RQ3 requires comparing across providers under matched reasoning configurations---a requirement that surfaces a subtle methodological issue: the GPT-5 model family reasons by default at the API level, while Claude models do not. Initial evaluations that do not correct for this asymmetry risk misattributing a reasoning-configuration effect to a capability gap.

Our contributions are three. First, we introduce \textsc{EnterpriseMem-Bench}, a programmatically constructed multi-turn Text-to-SQL benchmark with deterministic ground truth, memory-critical per-turn annotation, and a tier-based complexity structure spanning 300 sessions and 1,400 turns across three enterprise domains. Second, we report a systematic empirical study across five frontier models, five memory conditions, three datasets, and reasoning-enabled Claude configurations---35,000 turn evaluations in total---yielding findings on universal stateless collapse, non-monotonic memory-architecture returns, a reproducible and model-specific regression on SEC EDGAR financial queries, and a mono-modal error distribution under reasoning. Third, we document a methodological observation about reasoning-configuration asymmetry across API providers and demonstrate its empirical consequences on cross-provider benchmarking.

\section{Related Work}

\subsection{Single-Turn Text-to-SQL Benchmarks}

The modern Text-to-SQL evaluation literature is anchored by two benchmarks. Spider \citep{yu2018spider} introduced cross-domain Text-to-SQL evaluation across 200 databases with complex schemas, establishing component-match and execution-accuracy metrics that remain standard. BIRD \citep{li2023bird} extended this to larger databases with real-world data characteristics, including noisy values and external-knowledge requirements, and established execution accuracy (EX\%) as the primary evaluation metric in contemporary literature. Both benchmarks are overwhelmingly single-turn: each question is self-contained and independently answerable. Frontier prompting methods---notably DIN-SQL \citep{pourreza2023dinsql}, which decomposes Text-to-SQL into sub-tasks and applies self-correction---have driven execution accuracy on BIRD past 70\%, contributing to a widely reported perception that Text-to-SQL is a largely solved problem for production deployment.

This perception is misleading because it does not reflect the structure of analyst workflows. Production enterprise SQL conversations are overwhelmingly sequences of related questions rather than independent queries. Strong single-turn performance does not imply strong multi-turn performance---an observation that has only recently begun to receive direct empirical attention. A comprehensive survey of Text-to-SQL evaluation \citep{hong2024survey} confirms that multi-turn settings remain understudied relative to their practical importance.

\subsection{Multi-Turn Text-to-SQL}

Several prior benchmarks address multi-turn Text-to-SQL, but none treat memory architecture as an experimental variable. SParC \citep{yu2019sparc} and CoSQL \citep{yu2019cosql} introduced early conversational SQL datasets with user-system dialogue structures; their focus was on coreference and ellipsis rather than on explicit memory-system evaluation. MMSQL \citep{guo2024mmsql} studies multi-turn question types (clarification, confirmation, correction) but evaluates end-to-end accuracy rather than isolating memory effects. BIRD-INTERACT \citep{huo2025birdinteract} is the closest prior work: it extends BIRD to a multi-turn interactive setting with a function-driven user simulator and CRUD-spanning task suite, and confirms empirically that single-turn evaluation overstates realistic performance. It does not, however, define or ablate memory architectures---memory is implicit in the dialogue context the model receives, not a design variable under study.

Table~\ref{tab:benchmark_comparison} summarizes these benchmarks along four axes that motivate our work.

\begin{table*}[t]
\centering\footnotesize
\setlength{\tabcolsep}{6pt}
\begin{tabular}{lcccc}
\toprule
\textbf{Benchmark} & \textbf{Multi-turn} & \textbf{Memory-critical annotation} & \textbf{Memory-arch ablation} & \textbf{Programmatic GT} \\
\midrule
Spider & No & --- & --- & Human-authored \\
BIRD & No & --- & --- & Human-authored \\
SParC / CoSQL & Yes & No & No & Human-authored \\
MMSQL & Yes & No & No & Human-authored \\
BIRD-INTERACT & Yes & No & No & Mixed \\
\textsc{EnterpriseMem-Bench} (ours) & Yes & Yes (per turn) & Yes (three-way) & Yes \\
\bottomrule
\end{tabular}
\caption{Comparison of multi-turn Text-to-SQL benchmarks along axes relevant to memory-architecture study. ``Memory-critical annotation'' indicates whether the benchmark labels which turns require information from prior turns. ``Memory-arch ablation'' indicates whether the benchmark supports isolating contributions of distinct memory components.}
\label{tab:benchmark_comparison}
\end{table*}

\subsection{Memory Architectures for LLM Agents}

A parallel body of work studies memory architectures for LLM-based agents in general, without a Text-to-SQL focus. Multi-layer memory architectures \citep{zhang2024memorysurvey} decompose agent memory into working memory (short-term conversational state), episodic memory (retrievable past interactions), and semantic memory (structured knowledge or hints)---a decomposition we operationalize for Text-to-SQL in this work. MemGPT \citep{packer2023memgpt} frames LLM context windows as OS-like virtual memory with hierarchical tiers and explicit paging. A-MEM \citep{xu2025amem} introduces agentic memory with Zettelkasten-inspired note construction and dynamic link generation over an embedding store. Memory evaluation on long-conversation tasks \citep{maharana2024locomo} provides one of the principal benchmarks for measuring conversational memory; its question-answering protocol motivates our use of execution accuracy as a programmatic correctness signal. Reasoning models build on chain-of-thought work \citep{wei2022cot} demonstrating that intermediate-step reasoning improves complex task performance.

Text-to-SQL provides a particularly clean testbed for memory evaluation precisely because correctness is programmatically checkable: execution accuracy against ground-truth SQL is binary, deterministic, and requires no human judgment. Memory-competency evaluation on general conversational data evaluates capabilities in isolation rather than measuring memory's contribution to a downstream task with independently-verifiable correctness.

\subsection{Positioning}

This work is the first, to our knowledge, to treat memory-architecture type and complexity as the primary experimental variable in multi-turn Text-to-SQL. We build on BIRD-INTERACT's empirical observation that multi-turn performance lags single-turn performance, but we extend the question from ``does multi-turn evaluation matter?'' to ``which memory architecture is required, and what does each component contribute?'' The three-way ablation we introduce permits effects that in prior work were confounded---window size, retrieval, and augmentation---to be isolated as independent contributions. The per-turn memory-critical annotation we introduce permits a benchmark-intrinsic control that prior multi-turn benchmarks do not provide.

\section{Problem Formulation}

This section formalizes the concepts on which the remainder of the paper depends. We introduce them once, in isolation, before any experimental methodology or results depend on them.

\subsection{Multi-Turn Text-to-SQL}

Let $D$ denote a relational database with a fixed schema $S$. A natural-language question $q$ is mapped by a Text-to-SQL system to a SQL query $Q$; the query is executed against $D$, yielding a result set $R(Q, D)$. A ground-truth query $Q^*$ is said to be semantically equivalent to $Q$ if $R(Q, D)$ and $R(Q^*, D)$ are equal as unordered multisets. Execution accuracy on a dataset of $(q, Q^*)$ pairs is the fraction of system-generated $Q$ for which this equivalence holds.

A multi-turn Text-to-SQL interaction extends this formulation to a sequence. Rather than a single question, the system receives a sequence $q_1, q_2, \ldots, q_T$ of user questions issued in a single conversational session. At each turn $t$, the system must produce $Q_t$ given $q_t$ and, optionally, information from prior turns $1..t-1$. The crucial property that distinguishes multi-turn Text-to-SQL from concatenated single-turn Text-to-SQL is that some $q_t$ are not self-contained: they contain anaphoric references (e.g., pronouns, definite descriptions) that cannot be resolved without information from a prior turn.

\subsection{Sessions, Turns, and Memory-Critical Turns}

We define three benchmark-intrinsic concepts:

\paragraph{Session.} A session is a single conversational interaction consisting of an ordered sequence of turns that share a common topic and entity set. Sessions are independent of one another: information introduced in session $s_i$ is not available in session $s_j$ for $i \neq j$. A session is identified by a session identifier of the form \texttt{PREFIX-TIER-INDEX} (for example, \texttt{SEC-3-012} denotes the twelfth tier-3 session in the SEC EDGAR dataset).

\paragraph{Turn.} A turn is a single (question, ground-truth SQL) pair within a session, indexed sequentially from $t = 1$ to $t = T_s$ where $T_s$ is the session length. Each turn carries, in addition to the question and ground-truth SQL, a small structured payload: a \texttt{state\_updates} field recording key-value information introduced by the turn, a \texttt{referenced\_context\_keys} field listing the keys required to answer the turn, and a \texttt{memory\_benefit\_expected} flag described below.

\paragraph{Memory-critical turn.} A turn $t$ within a session is memory-critical if and only if answering it correctly requires information introduced in an earlier turn $t' < t$ within the same session. Formally, $t$ is memory-critical if the set of \texttt{referenced\_context\_keys} at $t$ is non-empty and every key in that set was bound by a \texttt{state\_update} at some earlier turn of the same session. Memory-criticality is a property of the benchmark---it is set deterministically during session construction and is independent of the model being evaluated.

By construction, every session's first turn ($t = 1$) is non-memory-critical: there is no prior turn from which information could be required. Whether subsequent turns are memory-critical depends on the session's tier. In tier-1 single-turn sessions, the session consists only of $t = 1$ and no memory-critical turns exist. In tier-2 and tier-3 multi-turn sessions, most or all non-first turns are memory-critical by design.

\paragraph{Example.} Consider the tier-2 SEC EDGAR session \texttt{SEC-2-001}, which begins: ``Show Apple revenue for 2022 and 2023.''---``Which year was higher?''---``Now focus on the newer year. What was net income?'' The first turn is non-memory-critical: it contains all information needed for generation (ticker = AAPL, metric = revenue, years = \{2022, 2023\}). The second turn has \texttt{referenced\_context\_keys} = [ticker, metric, older\_year, newer\_year]; all four keys were bound by the first turn's \texttt{state\_updates}. The second turn is therefore memory-critical. The third turn has \texttt{referenced\_context\_keys} = [ticker, newer\_year]; both keys are available from prior turns (ticker from $t = 1$, newer\_year from $t = 1$). The third turn is memory-critical.

Across \textsc{EnterpriseMem-Bench}, 300 sessions contain 300 non-memory-critical first turns (one per session) and 1,100 memory-critical subsequent turns---a memory-critical proportion of 78.57\%. This proportion is a deterministic consequence of benchmark design, not an empirical measurement; it follows from the session construction procedure.

\subsection{Execution Accuracy and Memory Benefit Score}

\paragraph{Execution Accuracy (EX\%).} For a turn with generated query $Q$ and ground-truth query $Q^*$, we define execution accuracy by executing both queries against the source database and comparing the result sets. A turn is judged correct if and only if: (i) both $Q$ and $Q^*$ execute without error, and (ii) the result sets match exactly after row-sorting and value-stringification. Execution accuracy on a set of turns is the fraction of correct turns. This definition is stricter than component-level match metrics (which check SQL structural similarity) but is standard in contemporary Text-to-SQL literature.

\paragraph{Memory Benefit Score (MBS).} To quantify the per-turn contribution of a memory condition $c$ relative to the stateless baseline $A$, we define $\mathrm{MBS}_c(t) = \mathrm{EX\%}_c(t) - \mathrm{EX\%}_A(t)$, reported in percentage points and computed at each turn depth $t$. By construction, $\mathrm{MBS}_c(t = 1) \approx 0$ on any well-designed memory benchmark because first turns are non-memory-critical and the stateless baseline is expected to perform equivalently to any memory condition. We find this property to hold empirically, serving as a sanity check on the benchmark design.

Two properties of MBS are useful in diagnostic analysis. First, on benchmarks where memory-critical turns increase in frequency with turn depth, MBS exhibits a characteristic ``hook curve'' shape: near-zero at $t = 1$, rising rapidly through $t = 2$ and $t = 3$, plateauing at higher depths. Second, because MBS is a difference of two accuracy values measured on the same set of turns, it is not sensitive to base-rate turn difficulty---which cancels---but is sensitive to the interaction between memory availability and turn difficulty.

\section{EnterpriseMem-Bench}

\textsc{EnterpriseMem-Bench} is a multi-turn Text-to-SQL benchmark designed specifically to support memory-architecture evaluation. This section describes its design principles, construction pipeline, and final statistics.

\subsection{Design Principles}

Four design principles govern the benchmark's construction. First, \textbf{real data}: all three datasets are built from publicly available databases with real content---real Czech bank loan records, real SEC EDGAR financial filings, real Northwind sales transactions. This distinguishes our benchmark from synthetic-data benchmarks whose ground-truth SQL might not survive realistic value distributions. Second, \textbf{deterministic ground truth}: every question and its corresponding ground-truth SQL is authored programmatically in Python. No large language model is involved in benchmark construction: question text, SQL templates, and expected result summaries are all written as code. Third, \textbf{memory-critical annotation}: every turn carries a deterministic flag, set during session construction, indicating whether the turn is memory-critical. This annotation is the benchmark's central instrumentation: it permits stateless-baseline collapse to be measured on exactly the turns where memory is necessary, separately from turns where memory is irrelevant. Fourth, \textbf{reproducibility}: all 300 sessions are constructed from public source databases by a single deterministic script.

\subsection{Source Databases}

We select three databases providing complementary enterprise-adjacent domains and schema characteristics (Table~\ref{tab:datasets}).

\begin{table}[h]
\centering\footnotesize
\setlength{\tabcolsep}{4pt}
\begin{tabular}{lrrrr}
\toprule
\textbf{Dataset} & \textbf{Sess.} & \textbf{Turns} & \textbf{Domain} & \textbf{Tbls} \\
\midrule
BIRD Financial & 80 & 334 & Czech banking & 8 \\
SEC EDGAR & 120 & 599 & F500 fin. (20 cos.) & 3 \\
Northwind & 100 & 467 & Enterprise sales & 13 \\
\midrule
\textbf{Total} & \textbf{300} & \textbf{1{,}400} & --- & --- \\
\bottomrule
\end{tabular}
\caption{Source databases. Total evaluations $=1{,}400 \times 5 \times 5 = 35{,}000$.}
\label{tab:datasets}
\end{table}

The three datasets are complementary in schema structure, which is intentional. BIRD Financial is a multi-entity relational schema requiring nontrivial joins (loan $\rightarrow$ account $\rightarrow$ disp[type=OWNER] $\rightarrow$ client $\rightarrow$ district). SEC EDGAR is a single-entity-dominant schema where most queries reduce to a WHERE clause on the \texttt{annual\_financials} table. Northwind sits between these: multiple tables, but most analyst queries join through a small number of paths. Varying schema structure across datasets allows us to study whether memory-architecture effects generalize across schema styles.

\paragraph{BIRD Financial.} The BIRD Financial subset uses the Czech banking \texttt{financial.sqlite} database with 8 tables. The dominant join structure we use is the ownership chain loan $\rightarrow$ account $\rightarrow$ disp (filtered to \texttt{type='OWNER'}) $\rightarrow$ client $\rightarrow$ district. Sessions are parameterized by (region, loan\_status) combinations drawn from this join. Region names are stored in \texttt{district.A3}; loan status values are raw BIRD codes A, B, C, D (loan lifecycle states).

\paragraph{SEC EDGAR.} The SEC EDGAR subset covers 20 Fortune 500 companies: AAPL, MSFT, JPM, NVDA, GOOGL, AMZN, META, TSLA, V, XOM, and 10 others. We construct sessions from the \texttt{annual\_financials} table (ticker, fiscal\_year, metric, value\_billions) containing 1,897 annual records spanning 2009--2026. The five metrics used are revenue, net\_income, cash, total\_assets, and total\_liabilities; session construction requires each selected (ticker, metric) pair to have data for at least two fiscal years.

\paragraph{Northwind.} The Northwind subset uses the standard Northwind schema with 93 customers and 16,282 orders. Sales totals are computed consistently via \texttt{UnitPrice * Quantity * (1 - Discount)} on the \texttt{[Order Details]} table; this formula is surfaced as a semantic-memory hint to allow memory-conditioned models to benefit from explicit join/formula guidance.

\subsection{Session Generation Pipeline}

Sessions are generated by a single Python script (\texttt{build\_sessions.py}, 1,451 lines) whose three-phase structure is: (1)~\textbf{Profile extraction}---for each dataset, the script issues deterministic SQL queries against the live source database to extract eligible entity profiles; (2)~\textbf{Template instantiation}---for each of three tiers, a hand-authored turn-sequence template is instantiated with one profile per session; templates specify question text, parameterized GT SQL, a fallback SQL for a different profile, \texttt{requires\_prior\_context} boolean, \texttt{referenced\_context\_keys} list, and \texttt{state\_updates} dictionary; (3)~\textbf{Verification}---the script asserts post-conditions on the generated benchmark: 300 sessions total, 1,400 turns, tier distribution \{T1: 90, T2: 120, T3: 90\}, source distribution \{Northwind: 100, SEC: 120, BIRD: 80\}. Session construction fails fast if these invariants do not hold. Ground-truth SQL for each turn is implicitly verified by being executable against the source database at agent-evaluation time; turns whose GT SQL does not return a valid result would surface as universal failures in \S\ref{sec:results} and were not observed.

\subsection{Ground-Truth Authoring}

Ground-truth SQL is authored by the benchmark author (first author) directly in the session-generation script. Each SQL query is hand-written against the source schema with specific choices about canonical forms: date filtering via \texttt{strftime('\%Y', ...)} on text-stored dates, explicit type handling for \texttt{value\_billions}, consistent join aliases, and \texttt{ROUND(...)} application for monetary aggregates. The same author-chosen canonical forms are applied across all sessions within a dataset, so that ground-truth SQL does not vary stylistically across turns.

This authoring model has a key implication: we do not use inter-annotator agreement as a benchmark-quality signal, because there is no annotator---there are programmatic templates. A different authoring choice (e.g., using arithmetic date extraction rather than \texttt{strftime}) would produce different SQL text but the same result set, and our execution-accuracy metric is indifferent to such text differences. Benchmark quality is instead defensible on the basis of: (1)~the SQL executes and returns the expected result type against the live database during template authoring; and (2)~all 300 sessions are reproducible from the public script.

\subsection{Memory-Critical Annotation}

The memory-critical property is set at template-authoring time by a deterministic rule: a turn is flagged memory-critical iff its natural-language question contains an anaphoric reference (pronouns like ``them''; deictic expressions like ``there,'' ``that year''; ellipsis relying on prior entities) that cannot be resolved without a prior turn's state. For each such turn the author also lists the specific state keys required in \texttt{referenced\_context\_keys} (e.g., \texttt{["customer\_id", "focus\_year"]}). Setting these fields during authoring---rather than inferring them from question text at evaluation time---makes benchmark structure visible and auditable.

\subsection{Session Tiers}

Sessions are partitioned into three tiers by conversational depth (Table~\ref{tab:tiers}). The tier structure is designed to provide (a)~a control tier where memory is irrelevant (T1), (b)~a short-conversation tier typical of quick analyst queries (T2), and (c)~an extended-session tier where memory benefit accumulates with depth (T3).

\begin{table}[h]
\centering\footnotesize
\setlength{\tabcolsep}{4pt}
\begin{tabular}{lrr l}
\toprule
\textbf{Tier} & \textbf{Sess.} & \textbf{Turns} & \textbf{Description} \\
\midrule
T1 -- Single & 90 & 90 & Control; no memory \\
T2 -- 3--5 turns & 120 & 480 & Pronouns, carry-forward \\
T3 -- 6--10 turns & 90 & 830 & Full analyst sessions \\
\midrule
\textbf{Total} & \textbf{300} & \textbf{1{,}400} & --- \\
\bottomrule
\end{tabular}
\caption{Session distribution by tier. Within T3, 20 sessions (7 Northwind, 9 SEC, 4 BIRD) have 10 turns; the remainder have 9.}
\label{tab:tiers}
\end{table}

The memory-critical turn proportion across \textsc{EnterpriseMem-Bench} is 1,100 of 1,400 = 78.57\% (every session's first turn is non-critical; every subsequent turn is critical by design). Per-tier breakdown: T1 = 0\%, T2 = 75.0\%, T3 = 88.2--90.0\% depending on dataset. The distribution is set deterministically by the template structure.

\subsection{Fallback-SQL Evaluation Design}

An unusual feature of \textsc{EnterpriseMem-Bench} is the inclusion of a \texttt{fallback\_sql} field on every memory-critical turn. The fallback SQL is the ground-truth SQL that would be correct if the model retrieved a different profile from a neighboring session---specifically, if it cross-contaminated context from the next session's profile. The fallback-SQL mechanism enables a finer-grained diagnosis of memory failures than execution-accuracy alone: a turn whose generated SQL matches the fallback profile's expected result set is evidence of cross-session memory contamination, distinct from a turn whose generated SQL matches neither ground truth nor fallback (evidence of pure reasoning failure).

\section{Experimental Design}

\subsection{Research Questions, Operationalized}

Each research question from \S1 corresponds to a specific experimental contrast: \textbf{RQ1} compares Condition A (stateless) against Condition B (two-turn working memory) across all 1,400 turns for each of the five models; \textbf{RQ2} compares Conditions B, B\_wm5, C, and D pairwise using the three-way ablation structure---the key comparisons are B\_wm5 $-$ B (window-size effect), C $-$ B\_wm5 (episodic effect), D $-$ C (semantic effect); \textbf{RQ3} compares the five models within each condition, with particular attention to the within-family comparison Sonnet 4.5 vs Sonnet 4.6 (generational effect), the cross-family comparison Claude vs GPT (capability effect), and the reasoning-mode comparison across Claude configurations (methodology effect).

\subsection{Memory Conditions and the Three-Way Ablation}
\label{sec:conditions}

\begin{figure*}[t]
\centering
\includegraphics[width=0.85\textwidth]{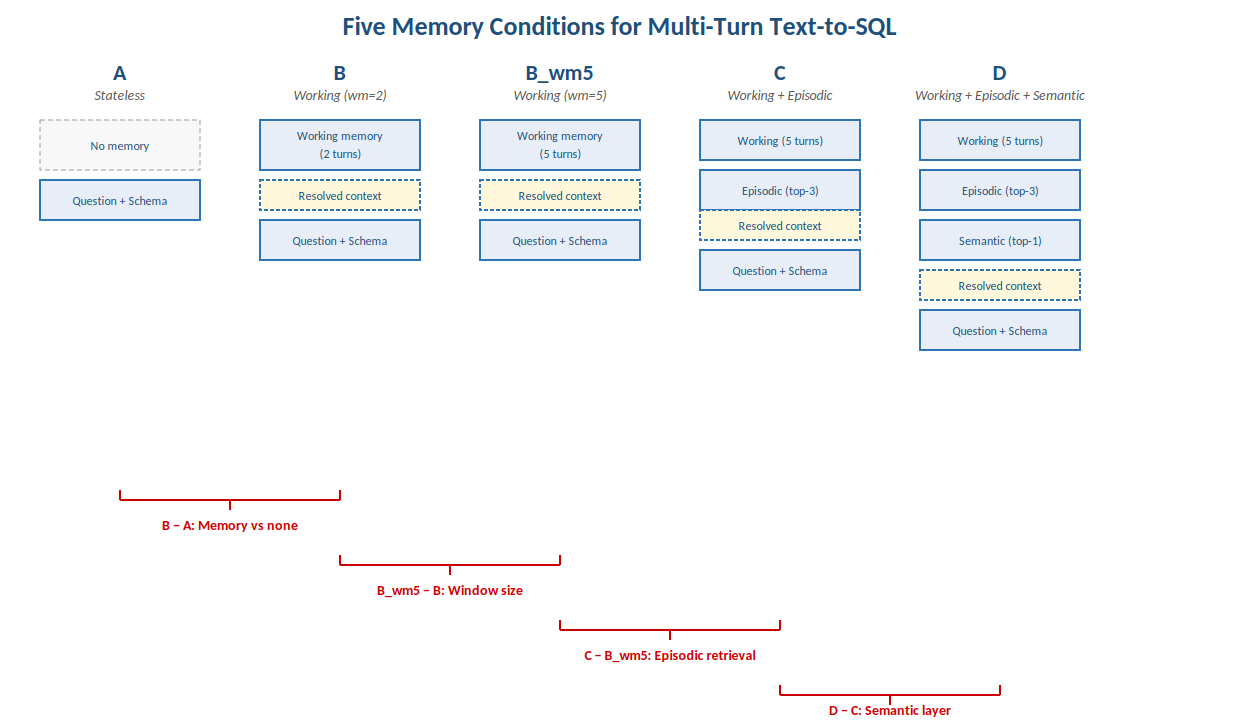}
\caption{Five memory conditions. Horizontal brackets at bottom indicate the three pairwise ablation comparisons: window-size ($\mathrm{B\_wm5}-\mathrm{B}$), episodic ($\mathrm{C}-\mathrm{B\_wm5}$), and semantic ($\mathrm{D}-\mathrm{C}$).}
\label{fig:arch}
\end{figure*}

We define five memory conditions of increasing architectural complexity (Figure~\ref{fig:arch}, Table~\ref{tab:conditions}). The central design choice is the inclusion of B\_wm5 as a distinct condition: it uses the same architecture as B (working memory only) but with a larger window (5 turns rather than 2). Without B\_wm5, a comparison C vs B would confound two effects---the increase in window size from 2 to 5 turns, and the addition of episodic retrieval. B\_wm5 separates these.

\begin{table}[h]
\centering\footnotesize
\setlength{\tabcolsep}{4pt}
\begin{tabular}{lcccc}
\toprule
\textbf{Cond.} & \textbf{Architecture} & \textbf{WM} & \textbf{Ep.\,$k$} & \textbf{Sem.\,$k$} \\
\midrule
A & Stateless & --- & --- & --- \\
B & Working & 2 & --- & --- \\
B\_wm5 & Working & 5 & --- & --- \\
C & WM + Ep. & 5 & 3 & --- \\
D & WM + Ep. + Sem. & 5 & 3 & 1 \\
\bottomrule
\end{tabular}
\caption{Five memory conditions with clean three-way ablation. B\_wm5 is the key design innovation: it separates window-size effects from episodic-retrieval effects, which would otherwise be confounded in any direct B vs C comparison.}
\label{tab:conditions}
\end{table}

\subsection{Models}

We evaluate five frontier LLMs across two providers (OpenAI, Anthropic) and multiple model generations. The selection is designed to support (a)~cross-family comparison (GPT vs Claude), (b)~within-family generational comparison (Sonnet 4.5 vs Sonnet 4.6), and (c)~within-family tier comparison (Sonnet vs Opus). API identifiers: \texttt{gpt-5-mini-2025-08-07}, \texttt{gpt-5.2-2025-12-11}, \texttt{claude-sonnet-4.5-20250929-v1-0}, \texttt{claude-sonnet-4-6-v1}, \texttt{claude-opus-4-6-v1}.

\subsection{Reasoning-Mode Methodology}

GPT-5 mini and GPT-5.2 are reasoning models that execute with built-in reasoning enabled by default and do not accept a temperature parameter. To ensure methodological consistency, all Claude models are evaluated with extended thinking enabled: Claude Sonnet 4.5 with \texttt{type="enabled"} and \texttt{budget\_tokens=5000}; Claude Sonnet 4.6 and Claude Opus 4.6 using adaptive thinking at \texttt{effort="medium"}. Generated SQL is extracted from text content blocks only; internal reasoning traces are not used for evaluation. The empirical and methodological implications of this parity choice---in particular, how much of the apparent Claude-GPT gap is attributable to reasoning configuration rather than capability---are discussed in \S\ref{sec:reasoning_compare}.

\subsection{Agent Architecture}

Evaluation is performed by a condition-aware SQL agent whose implementation (\texttt{base\_agent.py}) is released alongside the benchmark. The agent exposes a single \texttt{run\_turn} method that, given a session and a turn, performs four actions in parallel: context resolution (extracting named keys from working and episodic memory), working-memory retrieval (recent turns), episodic-memory retrieval (top-$k$ semantically similar past turns), and semantic-hint retrieval (top-$k$ schema hints). These four retrieval results are assembled into a prompt and sent to the model; the generated SQL is returned. After each turn, the agent persists the turn's payload (question, generated SQL, result summary, state updates) into working memory (if the condition uses it) and into episodic memory (if the condition uses it). Semantic memory is read-only and never written to during experiments.

\paragraph{Context Resolution.} A distinctive architectural feature is a context-resolution layer that operates independently of raw memory retrieval. Given a turn's \texttt{referenced\_context\_keys} (e.g., \texttt{["customer\_id", "focus\_year"]}), the agent walks working memory and, if keys remain unresolved, episodic memory, to extract specific key-value pairs. These resolved key-value pairs are then injected into the prompt as a separate \texttt{Resolved} structured context block, distinct from the raw conversation history. The system prompt instructs the model that resolved context values are authoritative and must be used exactly. This dual presentation---raw history plus extracted structured values---provides the model with redundant access to prior context through complementary modalities.

\subsection{Prompt Construction and Memory Stores}

All five models receive prompts constructed from a single template, parameterized by condition. The system prompt is identical across models and conditions (reproduced verbatim in Appendix~\ref{app:prompts}). Each dataset has hand-authored source-specific guidance (e.g., for Northwind: ``Sales totals come from \texttt{[Order Details]} using \texttt{UnitPrice * Quantity * (1 - Discount)}'') and per-question output requirements identical across all five models and all five conditions. This guidance creates a controlled evaluation environment in which differences across models reflect reasoning and memory utilization rather than schema familiarity or output-format ambiguity.

Working memory is implemented as a Redis-backed store keyed by session ID and cleared between sessions. Each session therefore starts with a fresh working memory regardless of condition. Episodic memory is implemented as a ChromaDB 1.0.9 collection using the in-built ONNX \texttt{all-MiniLM-L6-v2} embedding model. Episodic namespaces are unique per condition-run but the collection is shared across sessions within a single condition-run---this means episodic retrieval at session $s_i$ under condition $c$ can return episodes from sessions $s_1, \ldots, s_{i-1}$ evaluated under the same condition. This choice reflects production enterprise deployments where analyst episodic memory spans their entire work history rather than being session-isolated; it also introduces an empirical mechanism for cross-session retrieval effects that we analyze in \S\ref{sec:results}. Semantic memory is implemented as a static, pre-seeded ChromaDB namespace shared globally and never written to during experiments. It contains six hand-authored hints (two per dataset, covering join structure and computation formulas); retrieval top-$k$ is 1 at experiment time.

\subsection{Error Taxonomy and Classification}

Each turn's outcome is classified into one of six categories by the evaluation harness:

\begin{table}[h]
\centering\footnotesize
\setlength{\tabcolsep}{4pt}
\begin{tabular}{p{2.0cm}p{5cm}}
\toprule
\textbf{Category} & \textbf{Detection rule} \\
\midrule
Correct & Both generated and ground-truth SQL execute without error; result sets match after row-sort and value-stringification \\
Preamble & Generated output does not start with a SQL keyword but contains one later in the text \\
Syntax & Generated SQL produces an execution error containing syntax markers (``syntax error,'' parser-failure strings) \\
Runtime & Generated SQL produces an execution error without syntax markers (schema mismatch, type coercion failures) \\
Empty & Generated SQL string is empty \\
Wrong result & Generated SQL executes without error but result set does not match ground truth \\
\bottomrule
\end{tabular}
\caption{Error taxonomy. These six categories are mutually exclusive and collectively exhaustive. ``Schema error'' as a separate category is not reported because our programmatic classifier does not distinguish schema-mismatch runtime errors from other runtime errors; we report a single runtime category instead.}
\label{tab:error_taxonomy}
\end{table}

\subsection{Evaluation Inventory}

Total evaluations $= 1{,}400 \times 5 \times 5 = 35{,}000$ turn-evaluations. All experiments are single-run: each (turn, condition, model) triple is evaluated exactly once. Reasoning-mode Claude experiments were conducted in April 2026 using the configurations specified in \S5.4. Single-run evaluation is a deliberate methodological choice driven by evaluation cost and by the observation that our effect sizes are substantially larger than plausible single-run variance (stateless-collapse effects of 60+ percentage points; memory-architecture effects frequently exceeding 20 percentage points; EDGAR-regression effects of 17--33 percentage points). Reported point estimates should be interpreted as such; marginal comparisons warrant confirmation in future work with repeated runs.

\section{Results}
\label{sec:results}

\subsection{RQ1: Stateless Multi-Turn Collapse}

Table~\ref{tab:aggregate} reports aggregate EX\% across all 1{,}400 turns. Stateless (A) yields 15--19\% across all five models; working memory (B) recovers to 75--86\%.

\begin{table}[h]
\centering\footnotesize
\setlength{\tabcolsep}{2.5pt}
\begin{tabular}{lrrrrr}
\toprule
\textbf{Cond} & \textbf{S4.6} & \textbf{O4.6} & \textbf{S4.5} & \textbf{G5.2} & \textbf{G5m} \\
\midrule
A & 16.1 & 15.3 & 18.1 & 19.0 & 18.1 \\
B & 74.5 & 77.6 & 85.5 & 86.4 & 83.2 \\
B\_wm5 & 73.8 & 77.9 & 83.4 & 87.9 & 75.8 \\
C & 69.4 & 75.9 & 82.1 & 88.8 & 81.4 \\
D & 70.8 & 79.6 & 83.6 & 89.2 & 78.9 \\
\bottomrule
\end{tabular}
\caption{Aggregate EX\% (\%) across all datasets (1{,}400 turns). All Claude models with reasoning enabled. S4.6=Sonnet 4.6, O4.6=Opus 4.6, S4.5=Sonnet 4.5, G5.2=GPT-5.2, G5m=GPT-5 mini.}
\label{tab:aggregate}
\end{table}

\begin{figure}[h]
\centering
\includegraphics[width=\columnwidth]{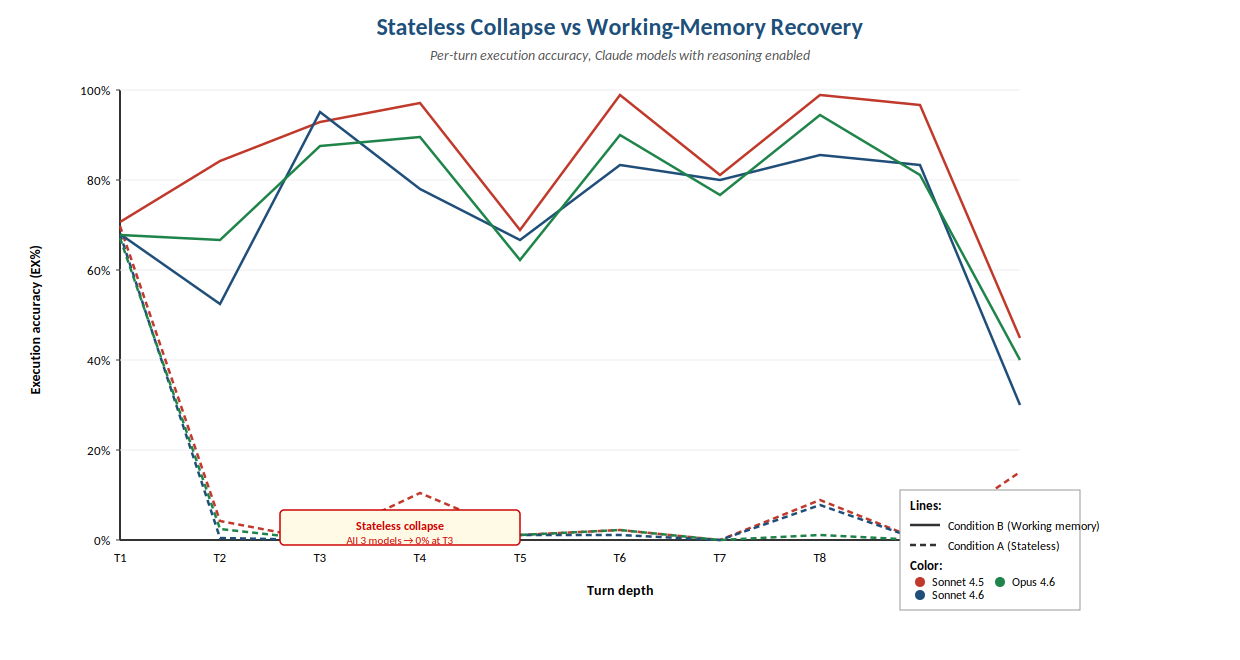}
\caption{Per-turn EX\%, Conditions A (dotted) and B (solid), Claude models with reasoning enabled. All three collapse from $\sim$68\% at T1 to exactly 0\% at T3 stateless. Working memory fully reverses the collapse.}
\label{fig:collapse}
\end{figure}

Aggregate numbers understate collapse severity because they mix T1 (memory irrelevant) with T2--T3 turns (memory essential). Figure~\ref{fig:collapse} shows the per-turn structure: Condition A drops to near-zero by Turn~3 and stays there; Condition B recovers to 60--95\%.

Turn-1 accuracy across all five models ranges from 67.0\% to 73.0\% at Condition A (Table~\ref{tab:perturn}), matching expected single-turn capability. The collapse occurs between T1 and T3 and is universal.

\noindent\textbf{Finding 1.} Stateless multi-turn Text-to-SQL collapses to 0\% EX\% at Turn 3 for all five models under reasoning (Table~\ref{tab:perturn}). The collapse is attributable to context loss: T1 accuracy is preserved (67.0--73.0\% across all five models), and a two-turn working window restores T3 EX\% to 87.6--100.0\% across all five models. The collapse is a memory failure, not a SQL-generation failure.

\begin{table*}[t]
\centering\footnotesize
\setlength{\tabcolsep}{4pt}
\begin{tabular}{llrrrrrrrrrr}
\toprule
\textbf{Model} & \textbf{Cond} & \textbf{T1} & \textbf{T2} & \textbf{T3} & \textbf{T4} & \textbf{T5} & \textbf{T6} & \textbf{T7} & \textbf{T8} & \textbf{T9} & \textbf{T10} \\
\midrule
Sonnet 4.5 & A & 69.7 & 4.3 & 0 & 10.5 & 1.1 & 2.2 & 0 & 8.9 & 0 & 15.0 \\
Sonnet 4.5 & B & 70.7 & 84.3 & 92.9 & 97.1 & 68.9 & 98.9 & 81.1 & 98.9 & 96.7 & 45.0 \\
Sonnet 4.6 & A & 68.0 & 0.5 & 0 & 5.2 & 1.1 & 1.1 & 0 & 7.8 & 0 & 5.0 \\
Sonnet 4.6 & B & 68.0 & 52.4 & 95.2 & 78.1 & 66.7 & 83.3 & 80.0 & 85.6 & 83.3 & 30.0 \\
Opus 4.6 & A & 67.0 & 2.4 & 0 & 1.9 & 1.1 & 2.2 & 0 & 1.1 & 0 & 0 \\
Opus 4.6 & B & 67.7 & 66.7 & 87.6 & 89.5 & 62.2 & 90.0 & 76.7 & 94.4 & 81.1 & 40.0 \\
\midrule
GPT-5 mini & A & 68.7 & 5.7 & 0 & 7.6 & 1.1 & 7.8 & 0 & 10.0 & 0 & 15.0 \\
GPT-5 mini & B & 68.3 & 76.2 & 100.0 & 91.4 & 70.0 & 91.1 & 83.3 & 91.1 & 96.7 & 45.0 \\
GPT-5.2 & A & 73.0 & 5.2 & 0 & 7.6 & 1.1 & 13.3 & 0 & 6.7 & 1.1 & 0 \\
GPT-5.2 & B & 73.0 & 90.0 & 100.0 & 96.2 & 70.0 & 96.7 & 81.1 & 100.0 & 74.4 & 50.0 \\
\bottomrule
\end{tabular}
\caption{Per-turn EX\% (\%) for Conditions A (stateless) and B (working memory, wm=2), all five models under reasoning (Claude with extended thinking enabled; GPT reasoning by default). The T10 column is based on $n=20$ sessions (7 Northwind, 9 SEC, 4 BIRD) and should be interpreted with caution for per-dataset analyses.}
\label{tab:perturn}
\end{table*}

Stateless EX\% reaches exactly 0\% at Turn 3 for all five models on memory-critical turns (Table~\ref{tab:perturn}). Turn-1 accuracy is competitive (67.0--73.0\% across all five models at Condition A), matching expected single-turn capability. The collapse occurs between Turn 1 and Turn 3 and is universal: every model tested, regardless of family, generation, or reasoning configuration, exhibits it. Working memory fully reverses the collapse: Condition B achieves 87.6--100.0\% at Turn 3 across all five models.

\subsection{RQ2: Memory-Component Contributions}

The MBS hook curve (Appendix~\ref{app:mbs}) confirms T1 values within 1pp of zero, validating benchmark construction. The three-way ablation (Table~\ref{tab:ablation}) reveals strong model- and dataset-dependence: window-size effects ($\mathrm{B\_wm5}-\mathrm{B}$) are small and usually negative; episodic effects ($\mathrm{C}-\mathrm{B\_wm5}$) range from $+4.5$ to $-12.6$; semantic effects ($\mathrm{D}-\mathrm{C}$) range from $+14.1$ (Opus 4.6 / EDGAR) to $-10.8$ (Opus 4.6 / BIRD). Net effects ($\mathrm{D}-\mathrm{B}$) span $+12.6$ to $-15.9$.

\begin{table}[h]
\centering\footnotesize
\setlength{\tabcolsep}{3pt}
\begin{tabular}{lrrrr}
\toprule
\textbf{Model / Dataset} & \textbf{Win} & \textbf{Ep} & \textbf{Sem} & \textbf{Net} \\
\midrule
S4.5 / BIRD & $-6.3$ & $-0.6$ & $+0.0$ & $-6.9$ \\
S4.5 / EDGAR & $+0.0$ & $-1.5$ & $+1.5$ & $+0.0$ \\
S4.5 / Northwind & $-1.7$ & $-1.5$ & $+2.6$ & $-0.6$ \\
S4.6 / BIRD & $-2.1$ & $+4.5$ & $+10.2$ & $+12.6$ \\
S4.6 / EDGAR & $-0.7$ & $-12.6$ & $-2.6$ & $-15.9$ \\
S4.6 / Northwind & $+0.2$ & $-0.2$ & $+0.2$ & $+0.2$ \\
O4.6 / BIRD & $-0.6$ & $-0.3$ & $-10.8$ & $-11.7$ \\
O4.6 / EDGAR & $+1.2$ & $-6.2$ & $+14.1$ & $+9.1$ \\
O4.6 / Northwind & $-0.5$ & $+2.4$ & $+0.9$ & $+2.8$ \\
\bottomrule
\end{tabular}
\caption{Three-way ablation. Effects in pp of EX\%. Win = $\mathrm{B\_wm5}-\mathrm{B}$; Ep = $\mathrm{C}-\mathrm{B\_wm5}$; Sem = $\mathrm{D}-\mathrm{C}$; Net = $\mathrm{D}-\mathrm{B}$.}
\label{tab:ablation}
\end{table}

\noindent\textbf{Finding 2.} Memory-architecture complexity does not monotonically improve EX\%. The largest single gain comes from Condition B (two-turn working memory), which recovers $60+$pp over stateless. Additional components produce effects from $+14$ to $-16$pp depending on model and dataset. No ablation is consistently positive across all combinations.

\subsubsection{When Does Memory Help?}

The model-and-dataset dependence of advanced memory components has a concrete deployment implication: memory-architecture choice is not a decision that can be made once for a deployed system. The same memory architecture that adds $+14$ percentage points for Opus 4.6 on EDGAR subtracts 11 percentage points for the same model on BIRD. A deployment serving mixed workloads must therefore either (a)~adopt the simpler memory architecture (Condition B) that is consistently positive across all models and datasets, or (b)~tune memory architecture per (model, dataset) pair.

\subsection{RQ3: Model-Family Differences}

Under reasoning parity, the two families are closer than prior non-matched evaluations suggest. At Condition B: GPT-5.2 $=86.4\%$, Sonnet 4.5 $=85.5\%$, GPT-5 mini $=83.2\%$, Opus 4.6 $=77.6\%$, Sonnet 4.6 $=74.5\%$. The GPT-5.2--Sonnet 4.5 gap is 0.9pp.

\subsubsection{The EDGAR Regression}

\begin{figure}[h]
\centering
\includegraphics[width=\columnwidth]{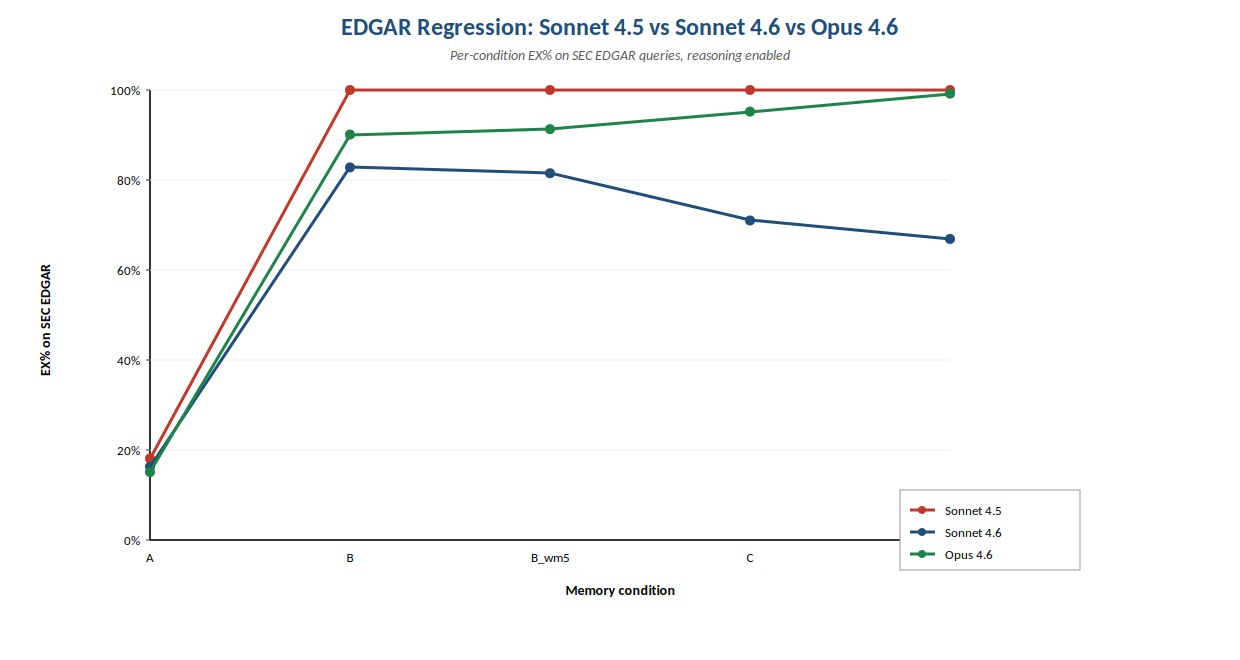}
\caption{EDGAR regression across conditions. Sonnet 4.6 underperforms Sonnet 4.5 by 17.2pp at B, 29.0 at C, 33.1 at D. Opus 4.6 reaches 99.2\% at D, matching Sonnet 4.5's ceiling.}
\label{fig:edgar}
\end{figure}

Sonnet 4.6 underperforms Sonnet 4.5 on SEC EDGAR by 17.2pp at B, 29.0 at C, and 33.1 at D. Opus 4.6 at D achieves 99.2\%, matching Sonnet 4.5's ceiling. The regression is specifically between Sonnet 4.5 and Sonnet 4.6 (generational, not family-wide), and dataset-specific: the same Sonnet-4.6 vs Sonnet-4.5 comparison on Northwind shows only 7.3pp gap; on BIRD, 5.1pp.

\noindent\textbf{Finding 3.} Claude Sonnet 4.6 underperforms Sonnet 4.5 by 17--33pp on SEC EDGAR across memory conditions. The regression is dataset-specific and generation-specific (Opus 4.6 does not exhibit it) and persists under reasoning, ruling out reasoning-configuration asymmetry as the explanation.

\subsubsection{Reasoning-Mode Comparison}
\label{sec:reasoning_compare}

Table~\ref{tab:reasoning} compares Claude EX\% under standard vs reasoning configurations against reasoning-by-default GPT baselines.

\begin{table}[h]
\centering\footnotesize
\setlength{\tabcolsep}{4pt}
\begin{tabular}{lrrr}
\toprule
\textbf{Model} & \textbf{Std} & \textbf{Reas.} & \textbf{$\Delta$ (pp)} \\
\midrule
Sonnet 4.5 & 79.8 & 85.5 & $+5.7$ \\
Sonnet 4.6 & 59.6 & 74.5 & $+14.9$ \\
Opus 4.6 & 58.7 & 77.6 & $+18.9$ \\
GPT-5.2 (default) & 86.4 & 86.4 & --- \\
GPT-5 mini (default) & 83.2 & 83.2 & --- \\
\bottomrule
\end{tabular}
\caption{Standard vs reasoning-mode comparison at Condition B. Reasoning closes 15--19pp of the gap for Sonnet 4.6 and Opus 4.6.}
\label{tab:reasoning}
\end{table}

Reasoning closes 15--19pp for Sonnet 4.6 and Opus 4.6. Sonnet 4.5 gains $+5.7$pp, reaching near-parity with GPT-5.2. We interpret this as evidence that a substantial fraction of the Claude-GPT gap in prior non-matched evaluations was a methodological artifact. The EDGAR regression persists under reasoning, reinforcing it as a model-specific effect.

\subsection{Error Taxonomy Under Reasoning}

Under reasoning-enabled evaluation, the error distribution for Claude models becomes sharply mono-modal. Table~\ref{tab:wr_rate} reports the dominant error type by condition for each Claude model, measured as the percentage of non-correct turns attributable to each category.

\begin{table}[h]
\centering\footnotesize
\setlength{\tabcolsep}{4pt}
\begin{tabular}{lrrr}
\toprule
\textbf{Cond.} & \textbf{S4.6 (WR\%)} & \textbf{O4.6 (WR\%)} & \textbf{S4.5 (WR\%)} \\
\midrule
A & 83.8 & 84.7 & 75.7 \\
B & 25.5 & 22.4 & 14.4 \\
B\_wm5 & 26.2 & 22.1 & 16.5 \\
C & 30.6 & 24.1 & 17.7 \\
D & 29.2 & 20.4 & 16.2 \\
\bottomrule
\end{tabular}
\caption{Wrong-result error rate (WR\%)---percentage of turns whose generated SQL executes cleanly but produces an incorrect result---across conditions and Claude models. Under all memory conditions, wrong-result errors dominate; preamble, syntax, and runtime errors are collectively negligible.}
\label{tab:wr_rate}
\end{table}

Specifically, under reasoning, Sonnet 4.6 and Opus 4.6 produce zero preamble errors across all 1{,}400 turns and all 5 conditions. Sonnet 4.5 produces 60 preamble errors total, all concentrated in Condition A, and concentrated on even-numbered turns (T2, T4, T6, T8) where the model is most likely to default to explanatory text when context is absent. Syntax errors are zero across all Claude models and all conditions; runtime errors are observed only 1--2 times across the entire Sonnet 4.5 evaluation; empty SQL never occurs. Every non-correct turn (outside the 60 Sonnet-4.5 preambles and handful of runtime errors) is a wrong-result error: syntactically valid, schema-correct SQL that executes cleanly and returns an incorrect result.

\noindent\textbf{Finding 4.} Under reasoning-enabled evaluation, Claude Text-to-SQL exhibits a mono-modal error distribution: preamble, syntax, runtime, and empty errors collapse to near-zero, and every remaining failure is a wrong-result error (syntactically valid SQL returning the wrong answer). This concentrates the remaining improvement opportunity in a single dimension---query logic selection---rather than the mixed error profile observed in prior non-matched evaluations of Claude Text-to-SQL.

\subsubsection{Preamble Errors: A Two-Mechanism Phenomenon}

Historical evaluations of Claude on SQL generation without reasoning have reported substantial preamble-error rates (approximately 26\% for Sonnet 4.5, 38\% for Sonnet 4.6, 45\% for Opus 4.6 at Condition A). Under reasoning, these collapse to 4.3\%, 0\%, and 0\% respectively. This observation suggests two independent mechanisms by which preamble errors can be eliminated: (i)~working memory injection, which provides the model with prior SQL-first responses as implicit format demonstrations; and (ii)~extended thinking, which appears to cause the model to deliberate on format before emitting tokens. Either mechanism alone suffices; the combination (Condition B + reasoning) eliminates preambles entirely for all three Claude models.

This two-mechanism interpretation has a practical corollary: Claude-family deployments for structured output that operate without reasoning should use working-memory injection as a format-stabilization technique, not merely as a context-carrying technique. The working-memory benefit on preamble elimination is independent of its benefit on answer correctness, and both are available from a minimal wm=2 configuration.

\section{Discussion}

\subsection{Synthesis Across Research Questions}

The four findings, taken together, recast a picture of multi-turn Text-to-SQL that differs substantively from single-turn evaluations. Stateless evaluation silently assumes models retain context; that assumption fails universally (Finding 1). Once working memory is provided, the additional-architecture question becomes a second-order decision conditional on (model, workload), with answers ranging from strongly positive to strongly negative (Finding 2). Cross-family comparisons distort when reasoning configurations are not matched, but within-family comparisons under matched reasoning can still reveal genuine and practically important generational regressions (Findings 3 and 4).

\subsection{Why Episodic and Semantic Underperform Expectations}

A natural expectation---that more memory architecture should yield more accuracy---does not survive empirical contact with the three-way ablation. The asymmetry demands a mechanistic account. Episodic retrieval presents episodes selected by embedding similarity to the current question, drawn from across the session history and, in our configuration, across prior sessions. Two failure modes are latent in this design. An episode retrieved from a different session may share surface similarity with the current question while involving different entities, yielding confidently-wrong SQL keyed on the wrong profile (cross-session contamination). An episode retrieved from earlier in the current session may surface state that has since been overwritten by a more recent turn, yielding stale context. Working memory, because it is restricted to the most recent $N$ turns of the current session, is structurally immune to both: it cannot retrieve across sessions, and it cannot surface state older than the window.

Semantic hints exhibit a parallel two-sidedness at a different scale. A hint that names a join path or an aggregation formula helps when the model would otherwise miss it---which explains gains like Opus 4.6 on EDGAR ($+14.1$pp) and Sonnet 4.6 on BIRD ($+10.2$pp). But when the model would have been correct and the hint contradicts or redirects its approach, the hint harms---which explains losses like Opus 4.6 on BIRD ($-10.8$pp). The mechanism is the same in both directions: retrieval biases the model toward the retrieved content, whether or not that content is correct for this turn.

\subsection{Implications for Enterprise Deployment}

Our findings yield three concrete deployment recommendations for production multi-turn Text-to-SQL systems.

\paragraph{1. Working memory is mandatory, not optional.} A stateless deployment will exhibit near-zero execution accuracy on memory-critical turns. A minimal two-turn working-memory window recovers the overwhelming majority of this gap. Memory is not a feature to ship later---it is load-bearing infrastructure.

\paragraph{2. Memory complexity should be justified empirically, not by default.} Adding episodic retrieval or semantic hints can help or hurt depending on model and workload. Deployments serving mixed workloads should default to the simpler architecture that is consistently positive, and add complexity only where empirical evaluation on the target workload demonstrates net benefit.

\paragraph{3. Cross-generation regressions are real and domain-specific.} Newer does not mean uniformly better within a model family. The Sonnet 4.5 $\rightarrow$ Sonnet 4.6 transition demonstrates a financial-domain regression of 17--33 percentage points that persists across memory configurations and reasoning modes. Deployers should evaluate on their target domains rather than relying on aggregate benchmark improvements.

\subsection{The Reasoning-Asymmetry Critique}

The methodology choice we describe in \S5.4 generalizes beyond this specific study. GPT-5 family models reason by default at the API level; Claude models do not. Evaluations that compare the two families without enabling extended thinking on the Claude side are comparing one model's full-capability output against another model's capability-restricted output. We measured this cost at 15--19 percentage points of aggregate EX\% for Sonnet 4.6 and Opus 4.6 under our benchmark conditions. We suspect similar effects are present in other published cross-provider comparisons and recommend that future benchmarks document reasoning configuration as a first-class methodological variable, on equal footing with temperature and decoding strategy.

\section{Conclusion}

We have presented \textsc{EnterpriseMem-Bench}, the first multi-turn Text-to-SQL benchmark designed specifically for memory-architecture evaluation, and reported a systematic 35{,}000-turn empirical study across five frontier models. Our four principal findings---universal stateless collapse, non-monotonic memory-architecture returns, the model-specific EDGAR regression, and mono-modal error distributions under reasoning---paint a picture of multi-turn SQL performance that differs substantively from what single-turn benchmarks suggest.

The work opens several directions. First, the EDGAR regression between Sonnet 4.5 and Sonnet 4.6 is an empirically clean signal of a specific capability tradeoff; its underlying mechanism is not known from our data and would benefit from focused follow-up evaluation. Second, the model- and dataset-dependent behavior of episodic and semantic memory suggests that memory-utilization-aware routing---selecting memory architecture per (model, workload) pair---may be a productive avenue for production deployment. Third, the reasoning-asymmetry methodological observation applies to cross-provider comparisons beyond Text-to-SQL and warrants explicit treatment in future benchmarking work. \textsc{EnterpriseMem-Bench}, the session-generation script, the agent implementation, and the evaluation harness are available to researchers upon request subject to institutional review.

\section*{Limitations}

\paragraph{Single-Run Evaluation.} All experiments are single-run per (turn, condition, model) triple. We do not report variance bars or statistical-significance tests on individual comparisons. The decision is driven by evaluation cost (5 models $\times$ 5 conditions $\times$ 1{,}400 turns $=$ 35{,}000 API calls per run) and justified by effect sizes that are substantially larger than plausible single-run variance: stateless-collapse effects exceed 60 percentage points; EDGAR-regression effects are 17--33 percentage points; memory-architecture effects are routinely 10+ percentage points. Nonetheless, reported point estimates should be interpreted as such, and marginal comparisons---for example, small differences in the ablation table---warrant confirmation in future work with repeated runs.

\paragraph{Three Datasets; Templated Questions.} \textsc{EnterpriseMem-Bench} comprises three source databases and uses hand-authored question templates instantiated with profile values, rather than natural analyst utterances collected from production systems. Templated question generation eliminates natural-language diversity as a confounding variable---all models face the same phrasing---but also limits the benchmark's coverage of real-world question styles. Expanding to additional schemas and to collected natural utterances is a clear direction for future work.

\paragraph{Medium Reasoning Only.} Claude Sonnet 4.6 and Opus 4.6 are evaluated at \texttt{effort="medium"} adaptive thinking; Sonnet 4.5 at \texttt{budget\_tokens=5000} extended thinking. We do not sweep reasoning budgets. Whether higher reasoning budgets would close the remaining gap between Sonnet 4.6 and Sonnet 4.5 on EDGAR---or between Claude and GPT in the aggregate---is an open question.

\paragraph{Single Benchmark Author.} Question templates and ground-truth SQL are authored by the first author alone. There is no annotator-agreement measurement because there are no annotators---the benchmark is programmatic. This is a methodological departure from human-annotated benchmarks and worth stating explicitly. Our defense: all ground-truth SQL is independently executable against the source databases, and authoring choices are visible in the released script. Still, a future community-reviewed version of the benchmark with multiple template authors would improve robustness.

\paragraph{SQLite Execution Environment.} All three source databases are packaged and queried as SQLite. This matches BIRD's evaluation convention and ensures cross-dataset comparability, but enterprise production Text-to-SQL systems typically run against more feature-rich SQL engines (PostgreSQL, Snowflake, BigQuery). Dialect differences---window functions, recursive CTEs, date-arithmetic syntax---could affect results on production dialects.

\paragraph{Reasoning Determinism.} Reasoning-enabled models are not strictly deterministic even at fixed parameters. Our single-run evaluation does not quantify reasoning-induced variance. Informal observation across small samples did not reveal high variance, but we do not present a systematic analysis.

\section*{Ethical Considerations}

The three source databases are publicly available: BIRD under its published license, SEC EDGAR as US government public-domain data, Northwind under an open-source license. The benchmark contains no human-subjects data and no personally identifiable information---session content is generated from public database records using hand-authored question templates. SEC EDGAR includes company-identifying data (e.g., ticker symbols of publicly listed firms), but only public corporate financial data; no individual identification is possible.

Our 35{,}000 turn-evaluations incurred non-trivial compute cost via commercial API calls; the environmental and monetary footprint of large-scale LLM benchmarking is a legitimate community concern, and we encourage future work to report such costs systematically. We further note that our methodology depends on commercial closed-weight API models (GPT-5 family, Claude family); this introduces a dependency on proprietary infrastructure that limits reproducibility for researchers without API access. We mitigate this partially by reporting model identifiers and configurations verbatim, but the underlying dependency remains.

Our deployment recommendations are stated for production multi-turn Text-to-SQL contexts; we caution against extrapolating them to high-stakes settings (e.g., medical record access, financial-decision-making) where additional auditability and access-control considerations apply. We are not aware of dual-use risks specific to the benchmark: its purpose is evaluating memory architectures for structured query generation. Underlying frontier LLMs are known to exhibit biases reflecting their training data, and these biases can propagate into generated SQL; deployers should evaluate for fairness on their target user populations before production use.

\section*{Acknowledgements}

The authors used Claude (Anthropic), a generative AI assistant, for substantive writing assistance during manuscript preparation. Specifically: drafting prose from author-supplied research findings; restructuring sections including reorganization and repetition reduction; paraphrasing for concision; formatting tables and figures from author-supplied data; and verifying bibliography entries against published sources. Some structural and framing decisions emerged from interactive discussion with the assistant; the authors retained final decision authority on all such decisions and are fully responsible for the correctness of methods, results, and writing. The research design, implementation, experimental execution, data collection, and findings---including the \texttt{build\_sessions.py} session generation script (1{,}451 lines), \texttt{base\_agent.py} condition-aware SQL agent, and evaluation harness---are entirely the work of the human authors. The assistant was not used to generate experimental code, run experiments, or produce results.

\bibliographystyle{acl_natbib}

\appendix

\section{Verbatim Prompts and Source Guidance}
\label{app:prompts}

\subsection{System Prompt (identical across models and conditions)}

\begin{quote}\small\itshape
``You are an expert SQLite analyst. Generate one valid SQLite query for the user's question. Return SQL only. No markdown, comments, or explanation. Use only the provided schema and context. If resolved structured context is provided, those values are authoritative and must be used exactly. Match the requested result shape exactly. If the question asks for a total, return a single aggregate value only, not supporting columns, not grouped rows, and not explanatory text.''
\end{quote}

\subsection{Northwind Source Guidance}

\begin{itemize}[leftmargin=*,itemsep=0pt]
\item Customer names are stored in \texttt{Customers.CompanyName}.
\item \texttt{Orders.CustomerID} joins to \texttt{Customers.CustomerID}.
\item Sales totals come from \texttt{[Order Details]} using \texttt{UnitPrice * Quantity * (1 - Discount)}.
\item For yearly sales questions, use \texttt{Orders.OrderDate}, not \texttt{Orders.ShippedDate}.
\item If the question asks for total sales for one customer and one year, return one row with one numeric aggregate column only.
\item For those total-sales questions, do not return \texttt{CompanyName}, \texttt{OrderID}, or per-order subtotals.
\item Do not GROUP BY unless the user explicitly asks for a breakdown.
\item Do not confuse customer company names with product names or category names.
\item If resolved context contains \texttt{customer\_id}, use that exact \texttt{CustomerID}.
\end{itemize}

\subsection{SEC EDGAR Source Guidance}

\begin{itemize}[leftmargin=*,itemsep=0pt]
\item \texttt{annual\_financials} joins to \texttt{companies} by \texttt{ticker}.
\item Company names live in \texttt{companies.company\_name}.
\item \texttt{value\_billions} is the normalized metric used in benchmark queries.
\item Metric values must use exact lowercase stored values: \texttt{revenue}, \texttt{net\_income}, \texttt{cash}, \texttt{total\_assets}, \texttt{total\_liabilities}.
\item If resolved context contains \texttt{ticker}, \texttt{metric}, or years, use them exactly.
\end{itemize}

\subsection{BIRD Source Guidance}

\begin{itemize}[leftmargin=*,itemsep=0pt]
\item Region names are in \texttt{district.A3}.
\item \texttt{loan} joins to \texttt{account}, then \texttt{disp} where \texttt{type='OWNER'}, then \texttt{client}, then \texttt{district}.
\item Loan status values are raw codes A, B, C, D.
\item If resolved context contains \texttt{region} or \texttt{loan\_status}, use them exactly.
\end{itemize}

\subsection{Semantic Memory Seed (Verbatim)}

\paragraph{Northwind.} (1)~``Orders joins to [Order Details] by OrderID and to Customers by CustomerID''; (2)~``Sales totals come from UnitPrice * Quantity * (1 - Discount) on [Order Details]''.

\paragraph{SEC EDGAR.} (1)~``annual\_financials contains ticker, fiscal\_year, metric, and value\_billions''; (2)~``revenue is stored as metric='revenue' and year comparisons use fiscal\_year''.

\paragraph{BIRD.} (1)~``loan joins to account by account\_id; account joins client through disp where type='OWNER'''; (2)~``district.A3 stores the region name, and loan.status uses the raw BIRD codes A, B, C, D''.

\subsection{Per-Condition Prompt Composition}

\begin{table}[h]
\centering\footnotesize
\setlength{\tabcolsep}{4pt}
\begin{tabular}{lccccc}
\toprule
\textbf{Block} & \textbf{A} & \textbf{B} & \textbf{B\_wm5} & \textbf{C} & \textbf{D} \\
\midrule
Schema + guidance & \checkmark & \checkmark & \checkmark & \checkmark & \checkmark \\
Working memory & --- & wm=2 & wm=5 & wm=5 & wm=5 \\
Episodic (top-3) & --- & --- & --- & \checkmark & \checkmark \\
Resolved context & --- & \checkmark & \checkmark & \checkmark & \checkmark \\
Semantic (top-1) & --- & --- & --- & --- & \checkmark \\
Question + format & \checkmark & \checkmark & \checkmark & \checkmark & \checkmark \\
\bottomrule
\end{tabular}
\caption{Per-condition prompt composition.}
\end{table}

\section{MBS Hook Curve}
\label{app:mbs}

\begin{figure}[h]
\centering
\includegraphics[width=\columnwidth]{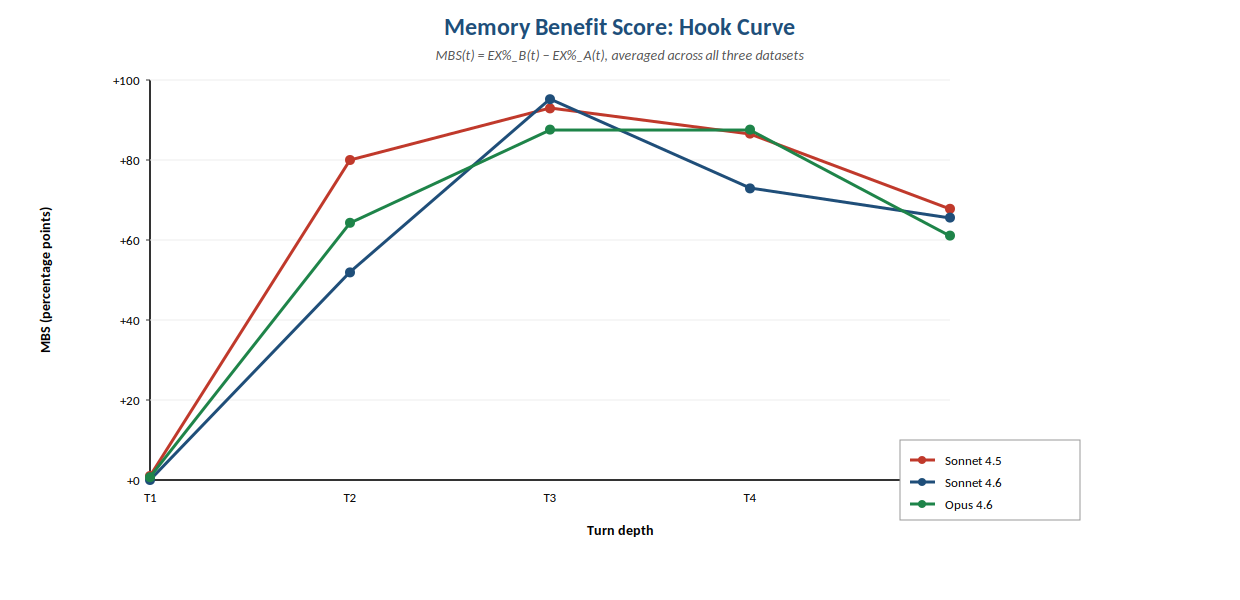}
\caption{MBS hook curve. $\textrm{MBS}(t) = \textrm{EX\%}_B(t) - \textrm{EX\%}_A(t)$, averaged across datasets. Near-zero T1, steep rise through T3, plateau thereafter.}
\end{figure}

\begin{table}[h]
\centering\footnotesize
\setlength{\tabcolsep}{4pt}
\begin{tabular}{lrrrrrr}
\toprule
\textbf{Model} & \textbf{T1} & \textbf{T2} & \textbf{T3} & \textbf{T4} & \textbf{T5} & \textbf{Avg} \\
\midrule
Sonnet 4.5 & $+1.0$ & $+80.0$ & $+92.9$ & $+86.6$ & $+67.8$ & $+65.9$ \\
Sonnet 4.6 & $0.0$ & $+51.9$ & $+95.2$ & $+72.9$ & $+65.6$ & $+57.4$ \\
Opus 4.6 & $+0.7$ & $+64.3$ & $+87.6$ & $+87.6$ & $+61.1$ & $+60.2$ \\
\bottomrule
\end{tabular}
\caption{MBS at each turn depth.}
\end{table}

\section{Per-Turn EX\% Detail}

\begin{table}[h]
\centering\footnotesize
\setlength{\tabcolsep}{2pt}
\begin{tabular}{llrrrrrr}
\toprule
\textbf{Model} & \textbf{C} & \textbf{T1} & \textbf{T2} & \textbf{T3} & \textbf{T5} & \textbf{T7} & \textbf{T9} \\
\midrule
Snt 4.5 & A & 69.7 & 4.3 & 0 & 1.1 & 0 & 0 \\
Snt 4.5 & B & 70.7 & 84.3 & 92.9 & 68.9 & 81.1 & 96.7 \\
Snt 4.6 & A & 68.0 & 0.5 & 0 & 1.1 & 0 & 0 \\
Snt 4.6 & B & 68.0 & 52.4 & 95.2 & 66.7 & 80.0 & 83.3 \\
Op 4.6 & A & 67.0 & 2.4 & 0 & 1.1 & 0 & 0 \\
Op 4.6 & B & 67.7 & 66.7 & 87.6 & 62.2 & 76.7 & 81.1 \\
\midrule
G5m & A & 68.7 & 5.7 & 0 & 1.1 & 0 & 0 \\
G5m & B & 68.3 & 76.2 & 100.0 & 70.0 & 83.3 & 96.7 \\
G5.2 & A & 73.0 & 5.2 & 0 & 1.1 & 0 & 1.1 \\
G5.2 & B & 73.0 & 90.0 & 100.0 & 70.0 & 81.1 & 74.4 \\
\bottomrule
\end{tabular}
\caption{Per-turn EX\% (\%), Conditions A and B, all five models under reasoning. Subset of turn depths shown; full table in Table~\ref{tab:perturn}. G5m=GPT-5 mini, G5.2=GPT-5.2.}
\end{table}

\section{Fallback-SQL Evaluation}

Every memory-critical turn carries a \texttt{fallback\_sql} field: the ground-truth SQL that would be correct if the model retrieved a neighboring session's profile. A turn whose generated SQL matches the fallback profile's expected result is evidence of cross-session memory contamination---distinct from a turn matching neither GT nor fallback (pure reasoning failure). All 250 failed Sonnet 4.6 turns at Condition B on EDGAR match neither GT nor fallback, ruling out retrieval contamination as the explanation of the regression.

\section{Example Session}

Session ID: \texttt{SEC-3-001} (Apple Inc., AAPL), 9 turns, tier T3:

\begin{itemize}[leftmargin=*,itemsep=2pt]
\item \textbf{Turn 1:} Show Apple revenue for 2022 and 2023.
\item \textbf{Turn 2:} Which year was higher?
\item \textbf{Turn 3:} Now show net income for those same two years.
\item \textbf{Turn 4:} Which year was higher there?
\item \textbf{Turn 5:} Show cash for those years.
\item \textbf{Turn 6:} Which year was higher for cash?
\item \textbf{Turn 7:} Show total assets for those years.
\item \textbf{Turn 8:} Which year was higher for total assets?
\item \textbf{Turn 9:} And total liabilities?
\end{itemize}

Turn 9 references ticker and years from Turn 1, eight turns prior. Only memory conditions with sufficient window depth or effective episodic retrieval can answer Turn 9 correctly.

\section{Standard-Mode vs Reasoning-Mode Comparison}
\label{app:reasoning_compare}

Table~\ref{tab:std_vs_reasoning} reports aggregate EX\% across all 1{,}400 turns at Condition B for Claude models under both configurations, alongside reasoning-by-default GPT baselines. This comparison is the basis for the reasoning-asymmetry claim in \S\ref{sec:results}.

\begin{table}[h]
\centering\footnotesize
\setlength{\tabcolsep}{4pt}
\begin{tabular}{lrrr}
\toprule
\textbf{Model} & \textbf{Standard} & \textbf{Reasoning} & \textbf{$\Delta$ (pp)} \\
\midrule
Claude Sonnet 4.5 & 79.8 & 85.5 & $+5.7$ \\
Claude Sonnet 4.6 & 59.6 & 74.5 & $+14.9$ \\
Claude Opus 4.6 & 58.7 & 77.6 & $+18.9$ \\
GPT-5 mini (default) & 83.2 & 83.2 & --- \\
GPT-5.2 (default) & 86.4 & 86.4 & --- \\
\bottomrule
\end{tabular}
\caption{Standard-mode vs reasoning-mode comparison at Condition B, aggregated across all three datasets. Standard-mode numbers reflect the pre-correction evaluation in which Claude models were invoked without extended thinking while GPT models reasoned by default.}
\label{tab:std_vs_reasoning}
\end{table}

\section{Reproducibility}

The following artifacts will be released upon publication:
\begin{itemize}[leftmargin=*,itemsep=0pt]
\item Session generation script (1{,}451 lines, Python) producing all 300 sessions with verifiable tier and source distributions
\item Condition-aware SQL agent implementing all five memory conditions against Redis + ChromaDB backend
\item Semantic memory seed (six hand-authored hints reproduced verbatim in supplementary materials)
\item Source databases: BIRD, SEC EDGAR, Northwind---all publicly available
\item Evaluation harness with six-category error taxonomy classifier
\item Raw per-turn results for all 35{,}000 evaluation instances
\end{itemize}

Model API identifiers:
\begin{itemize}[leftmargin=*,itemsep=0pt]
\item \texttt{gpt-5-mini-2025-08-07}
\item \texttt{gpt-5.2-2025-12-11}
\item \texttt{claude-sonnet-4.5-20250929-v1-0}
\item \texttt{claude-sonnet-4-6-v1}
\item \texttt{claude-opus-4-6-v1}
\end{itemize}
ChromaDB 1.0.9 with in-built ONNX all-MiniLM-L6-v2 embedder. Redis with default single-instance deployment.

\end{document}